\documentclass[10pt,twocolumn,letterpaper]{article}

\usepackage[final]{iccv} 
\usepackage{multirow}
\usepackage[accsupp]{axessibility}
\usepackage{pifont}
\definecolor{iccvblue}{rgb}{0.21,0.49,0.74}
\usepackage[pagebackref,breaklinks,colorlinks,allcolors=iccvblue]{hyperref}

\title{MPG-SAM 2: Adapting SAM 2 with Mask Priors and Global Context for Referring Video Object Segmentation}
\author{
{Fu Rong}$^{1}$, 
{Meng Lan}$^{2}$, 
{Qian Zhang}$^{3}$, 
{Lefei Zhang}$^{1}$\thanks{Corresponding author} \\
$^{1}$National Engineering Research Center for Multimedia Software, \\
School of Computer Science, Wuhan University \\
$^{2}$Hong Kong University of Science and Technology \quad 
$^{3}$Horizon Robotics
}
\begin{document}
\maketitle
\begin{abstract}
Referring video object segmentation (RVOS) aims to segment objects in a video according to textual descriptions, which requires the integration of multimodal information and temporal dynamics perception. The Segment Anything Model 2 (SAM 2) has shown great effectiveness across various video segmentation tasks. However, its application to offline RVOS is challenged by the translation of the text into effective prompts and a lack of global context awareness. In this paper, we propose a novel RVOS framework, termed MPG-SAM 2, to address these challenges. Specifically, MPG-SAM 2 employs a multimodal encoder to jointly encode video and textual features, generating semantically aligned video and text embeddings along with multimodal class tokens. A mask prior generator is devised to utilize the video embeddings and class tokens to create pseudo masks of target objects and global context. These masks are fed into the prompt encoder as dense prompts, along with multimodal class tokens as sparse prompts to generate accurate prompts for SAM 2. To provide the online SAM 2 with a global view, we propose a hierarchical global-historical aggregator, which allows SAM 2 to aggregate global and historical information of target objects at both pixel and object levels, enhancing the target representation and temporal consistency. Extensive experiments on several RVOS benchmarks demonstrate the superiority of MPG-SAM 2 and the effectiveness of the proposed modules. The code is available at \href{https://github.com/rongfu-dsb/MPG-SAM2}{https://github.com/rongfu-dsb/MPG-SAM2}.
\end{abstract}

\section{Introduction}
Referring video object segmentation (RVOS) \cite{Referformer, miao2023spectrum, luo2024soc, lan2024bidirectional} aims to segment the target objects in a video according to textual descriptions. This task, combining video segmentation \cite{lan2023learning} and language comprehension, necessitates not only proficiency in segmentation and inter-frame information propagation, as in traditional video object segmentation (VOS) \cite{oh2019video, cheng2022xmem, bekuzarov2023xmem++}, but also a robust understanding of the textual reference within the broader video context. The core challenges of RVOS thus lie in efficiently aligning multimodal information and maintaining temporal consistency. 

\begin{figure}[t]
    \centering
     \includegraphics[width=\linewidth]{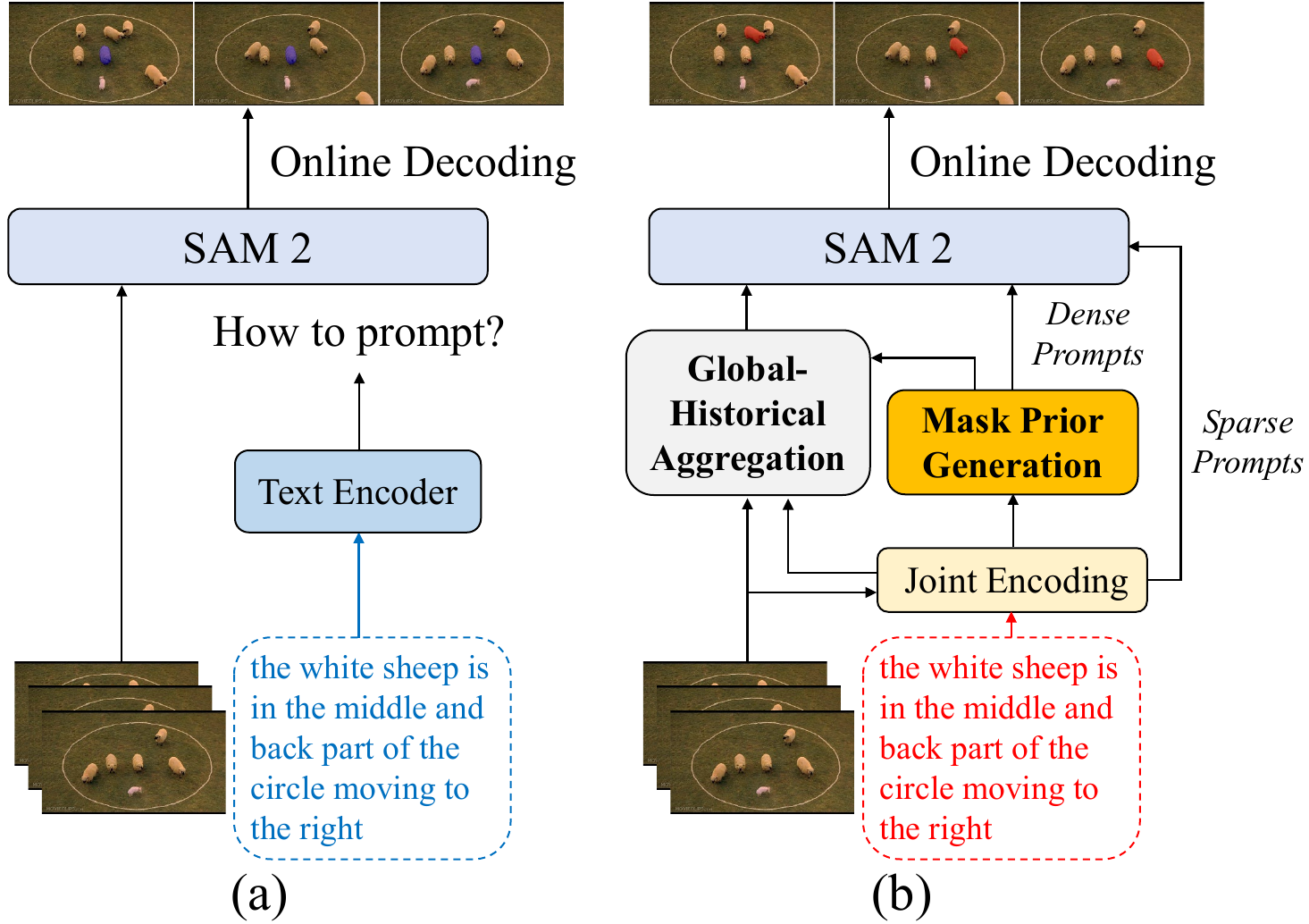}
     \caption{Comparison of two SAM 2 adaptations for RVOS. (a) vanilla SAM 2, (b) our MPG-SAM 2.}
     \label{fig1}
     \vspace{-4mm}
\end{figure}

 Recently, the Segment Anything Model (SAM) \cite{kirillov2023segment} and its variants \cite{xiong2024efficientsam, zhang2023faster, ke2024segment} have demonstrated significant improvements in efficiency and accuracy for promptable segmentation in images, leveraging robust segmentation abilities and interactive prompts. SAM 2 \cite{ravi2024sam} extends promptable segmentation from the image to the video domain by introducing a memory mechanism to enhance temporal consistency, achieving remarkable performance in VOS. However, its application to RVOS presents several challenges. 
 
 \textbf{First}, the inherent absence of textual prompts in SAM 2's architecture hinders the delivery of accurate prompts that align with the provided textual descriptions, as shown in \cref{fig1} (a). Although some efforts have explored this area, further improvement is still needed. For instance, RefSAM \cite{li2024refsamefficientlyadaptingsegmenting} projects text embedding into sparse and dense prompts for SAM, but the independently encoded text prompts may not fully capture visual semantics, limiting the effective use of SAM's segmentation capability. AL-Ref-SAM 2 \cite{huang2024unleashing} employs GPT-4 and Grounding DINO \cite{liu2023grounding} to translate textual information into a box prompt of the target object, but the multi-stage pipeline heavily depends on the upstream model's spatio-temporal reasoning ability, with substantial model parameters constraining deployment and inference efficiency of the model. Therefore, how to effectively align the vision-language features and provide accurate prompts to guide the decoding process is essential for adapting SAM 2 to RVOS. \textbf{Second}, the online-mode SAM 2 only has a historical view and cannot provide a global perspective for offline-mode RVOS, which may affect the global alignment of multimodal information and the temporal consistency of target objects. Consequently, effectively injecting global context information of the target objects into SAM 2 is crucial for RVOS.

To address these challenges, this paper introduces MPG-SAM 2, a novel end-to-end RVOS framework adapted from SAM 2. As illustrated in \cref{fig1} (b), \textit{our core innovation lies in generating precise prompts and global context through aligned video-text features for injection into SAM 2}. Specifically, we first employ an existing multimodal encoder to jointly encode input video and text, producing semantically aligned video and text embeddings, and multimodal class tokens. Then, we devise a novel mask prior generator, which leverages the video embeddings and multimodal class tokens to create the pseudo masks of target objects for each frame in the video, serving as dense prompts of SAM 2 that provide strong positional guidance for mask decoding. Additionally, following \cite{zhang2024evf}, we sent the multimodal class tokens to the prompt encoder as the sparse prompts after a multilayer perceptron (MLP). By combining the powerful dense and sparse prompts, accurate prompts are provided to the mask decoder for better performance.

To introduce the global context of target objects into SAM 2, we design a hierarchical global-historical aggregator that allows SAM 2 to aggregate global context and historical information of target objects at multiple levels before mask decoder. Here, the global context primarily comes from the global video feature generated by the mask prior generator. The aggregator consists of pixel and object-level fusion modules. In the pixel-level module, the current image feature interacts sequentially with the historical features in the memory of SAM 2 and the global context, thus enhancing the pixel-level target representation from various perspectives. Similarly, in the object-level module, the mask tokens aggregate target representation information from the global video feature and historical mask tokens in memory to generate the object tokens for the mask decoder. 

Experimental results on several RVOS benchmarks demonstrate the state-of-the-art performance of our model and the effectiveness of our proposed modules. The main contributions of this work can be summarized as follows:
\begin{itemize}
\item We propose a novel RVOS framework, MPG-SAM 2, adapted from SAM 2 by introducing mask prior-based dense prompt and multi-level global context fusions, achieving cutting-edge performance on several RVOS benchmarks.

\item We devise a mask prior generator that leverages the global video feature and multimodal class tokens to produce pseudo masks of target objects, providing the prior position cues as dense prompts for SAM 2 to enhance the mask decoding.

\item We develop a hierarchical global-historical aggregator that integrates global context and historical memory information of target objects into SAM 2 at both pixel and object levels. This module enables the online SAM 2 to have a global view and enhances the target representation and temporal consistency.

\end{itemize}

\begin{figure*}[t]
    \centering
     \includegraphics[width=\linewidth]{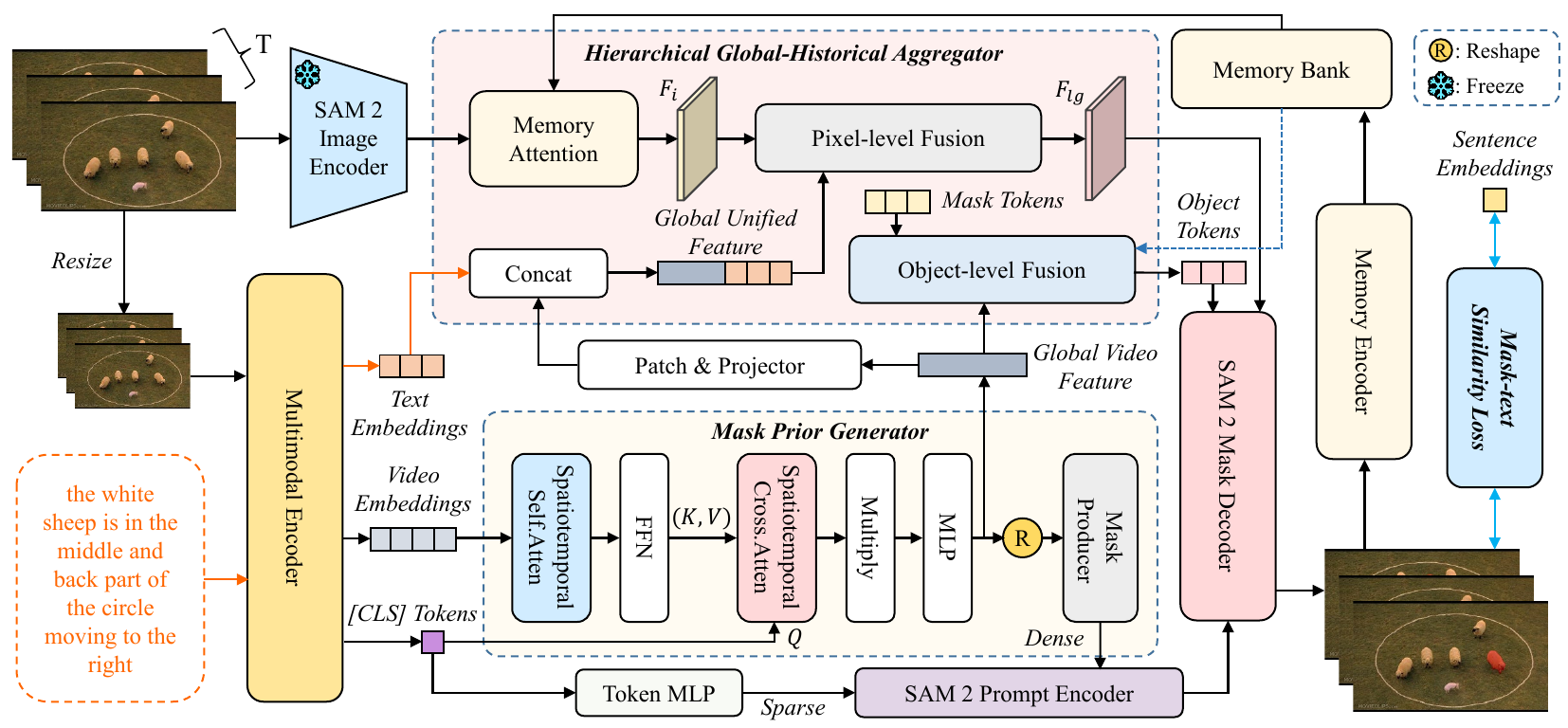}
     \caption{The overview of the proposed MPG-SAM 2. It mainly consists of four parts: the multimodal encoder, the mask prior generator, the hierarchical global-historical aggregator, and the SAM 2. The mask prior generator generates pseudo masks of target objects as the dense prompts of SAM 2 and produces the global video feature as the global context. The global-historical aggregator aggregates the target information from the global context and memory features to enhance the target representation of pixel-level image features and object-level object tokens. The mask-text similarity loss constrains the correlation between object masks and textual features.}
     \label{framework}
\end{figure*}

\section{Related work}

\noindent\textbf{Referring Video Object Segmentation.}
RVOS attracts extensive interest as it bridges visual and linguistic domains. Early methods, such as RefVOS \cite{bellver2020refvos}, consider RVOS as an expansion of referring image segmentation (RIS) to the video domain. URVOS \cite{Urvos} advances this by combining RIS and semi-supervised video object segmentation within a unified framework using attention mechanisms. In subsequent studies \cite{mcintosh2020visual, wang2019asymmetric, yofo, ding2022language, wu2022multi}, researchers emphasize cross-modal interaction \cite{fang2024unified, wu2025logiczsl}, further enhancing RVOS performance. However, despite notable gains, the computational expense and complexity of multi-stage pipelines limit practical feasibility. In response, query-based Transformer architectures \cite{lan2024language} offer efficient solutions with simplified yet robust frameworks. Notably, MTTR \cite{MTTR} and ReferFormer \cite{Referformer} pioneer the use of the DETR series \cite{DETR, zhu2020deformable} in RVOS, introducing novel multimodal interaction mechanisms. Recent methods \cite{luo2024soc, miao2023spectrum, tang2023temporal, lan2024bidirectional, pan2025semantic} refine these Transformer architectures through advanced temporal and multimodal feature integration techniques. For instance, SgMg \cite{miao2023spectrum} enhances ReferFormer by replacing dynamic convolution with a segmentation optimizer and uses spectral information to guide visual feature fusion. SOC \cite{luo2024soc} implements layered temporal modeling across video and object levels, achieving early multimodal fusion. More recently, Losh \cite{yuan2024losh} introduces a joint prediction network for short and long sentences, reducing the over-influence of action and relational cues in segmentation. VD-IT \cite{zhu2024exploring} explores the potential of text-to-video diffusion models in the RVOS task, leveraging the inherent rich semantics and coherent temporal correspondences of video generation models to ensure temporal instance consistency. To tackle longer sequences and complex scenarios in the MeViS \cite{ding2023mevis} dataset, DsHmp \cite{he2024decoupling} proposes a static text decoupling strategy, enhancing the temporal understanding of static and dynamic content at both frame and object levels, thereby capturing both short-term and long-term motion insights.

\noindent\textbf{Segment Anything Model.} SAM \cite{kirillov2023segment} is an interactive segmentation model capable of generating non-semantic masks based on various prompts. Trained on a large-scale dataset, SAM exhibits strong generalization across a wide range of common objects. Several variants \cite{xiong2024efficientsam, zhang2023faster, ke2024segment} have focused on improving the segmentation accuracy and computational efficiency of SAM. Moreover, SAM has found widespread application in various fields, including video tracking \cite{cheng2023segment}, remote sensing image interpretation \cite{wang2024samrs, rong2025customized}, and medical image processing \cite{yue2024surgicalsam}. Recently, SAM 2 \cite{ke2024segment} extends SAM to the video domain, achieving state-of-the-art performance. Despite its excellence in visual segmentation tasks using box, point, or mask prompts, SAM lacks language comprehension capabilities and cannot directly handle referring segmentation tasks. With the advancement of multimodal large language models (MLLMs), recent approaches \cite{lai2024lisa, xu2023u, rasheed2024glamm} utilize MLLMs to encode textual guidance for SAM segmentation. Specifically, LISA \cite{lai2024lisa} fine-tunes LLaVA \cite{liu2024visual} to generate multimodal features by extracting hidden embeddings based on specific prompts. u-LLaVA \cite{xu2023u} extends this approach to enable joint multi-task processing at region and pixel levels. GLaMM \cite{rasheed2024glamm} integrates the visual grounding task and incorporates language responses to provide multi-granular segmentation prompts to the model. Most recently, EVF-SAM \cite{zhang2024evf} introduces a lightweight pre-fusion architecture that employs joint visual-language encoding to generate high-quality textual prompts, leading to excellent segmentation performance.

\section{Method}
\subsection{Overview}
The overview of our proposed MPG-SAM 2 is illustrated in \cref{framework}. MPG-SAM 2 consists of four primary components: the multimodal encoder, the proposed mask prior generator, the devised hierarchical global-historical aggregator, and the SAM 2. Given a video sequence $\mathcal{V}=\left\{I_t\right\}_{t=1}^T$ with $T$ frames and its corresponding textual description $\mathcal{E}=\left\{e_l\right\}_{l=1}^L$ with $L$ words, the multimodal encoder first performs joint encoding independently across frames, extracting the multimodal [CLS] tokens, the video patch embeddings and the text embeddings. Simultaneously, the image encoder of SAM 2 independently extracts video frame features. The mask prior generator receives the video patch embeddings and multimodal [CLS] tokens, generates the global video feature and produces prior masks for each frame. These prior masks along with the multimodal [CLS] tokens, serve as the prompts for SAM 2. The hierarchical global-historical aggregator integrates the global video feature, text embeddings, and the historical mask features and tokens from SAM 2's memory to hierarchically enhance the target representations of pixel-level image features and object-level object tokens. Finally, SAM 2's decoder performs online decoding based on the provided prompts, object tokens, and current image features to obtain precise object masks for the current frame.

\subsection{Feature Extraction}
Aligning the visual-linguistic space is essential for RVOS. Following EVF-SAM \cite{zhang2024evf} and Shared-RIS \cite{yu2024simple}, we employ the unified BEiT-3 \cite{wang2023image} encoder to jointly encode video and language features. Each input frame $I_{m}\in{\mathbb{R}^{H_{m}\times{W_{m}}\times{3}}}$ is partitioned into non-overlapping patches $V_{p}\in{\mathbb{R}^{N_{v}\times({p^2}\times{3})}}$, which are subsequently projected into the feature space as $V_{p}\in{\mathbb{R}^{N_{v}\times{D}}}$. A visual class token $V_{cls}$ is prepended, followed by the addition of learnable positional embeddings $V_{pos}$, forming the final visual representation $V_{0}$. In parallel, the textual input of length $L$ is tokenized via XLMRobertaTokenizer \cite{conneau2019unsupervised}, producing $T_{tok}$, with a class token $T_{cls}$ and an end-of-sequence token $T_{end}$ appended. Positional embeddings $T_{pos}$ are incorporated to obtain the text representation $T_{0}\in{\mathbb{R}^{N_{l}\times{D}}}$, where $N_{l}=L+2$. The final joint representation $G_{0}$ is constructed by concatenating the visual embeddings $V_{0}$ and textual embeddings $T_{0}$. The process is formally expressed as:
\begin{equation}
\begin{aligned}
    &V_{0} = [V_{cls}, V_{p}] + V_{pos},\\
    &T_{0} = [T_{cls}, T_{tok}, T_{end}] + T_{pos},\\
    &G_{0} = [V_{0}; T_{0}] \in \mathbb{R}^{(N_{v}+N_{l}+1)\times{D}}.\\
\end{aligned}
\end{equation}

Following multimodal fusion through multiple attention blocks, the joint visual-text embedding is fed into modality-specific feed-forward neural network (FFN) for vision and text. We ultimately obtain the joint visual-text embeddings $G\in{\mathbb{R}^{T\times{(N_{v}+N_{l}+1)\times{D}}}}$ for the entire video, 
which are then decomposed into the multimodal [CLS] tokens $V_{cls}\in \mathbb{R}^{T\times{1}\times{D}}$, video patch embeddings $V \in \mathbb{R}^{T\times{N_{v}}\times{D}}$, and text embeddings $T \in{\mathbb{R}^{T\times{N_{l}}\times{D}}}$.

Meanwhile, the SAM 2 image encoder performs frame-independent encoding on each input video frame $I_{s}\in{\mathbb{R}^{{H_{s}}\times{W_{s}}\times{3}}}$. For the $i$-th frame, a set of hierarchical multi-scale features is extracted for mask generation, with the image feature of final layer  $F_{i}\in{\mathbb{R}^{\frac{H_s}{16}\times\frac{W_s}{16}\times{C}}}$ serving as the input for the subsequent decoding process, where $H_{s}$, $W_{s}$ and $C$ denote the height and width of the input image, and the channel of SAM 2 image encoder, respectively.

\subsection{Mask Prior Generator}

Despite the demonstrated effectiveness of semantic alignment between video patch embeddings $V$ and text embeddings $T$, we identify two critical limitations in the current framework: (1) SAM 2's video features $F$ inherently lack linguistic context, resulting in a semantic gap with text representations, and (2) the frame-agnostic characteristic of [CLS] tokens restricts their ability to model temporal dependencies across video sequences, often leading to segmentation artifacts including object misalignment and temporal inconsistency. To mitigate these limitations, we introduce a novel approach that generates frame-specific pseudo mask priors through the fusion of [CLS] tokens with language-enhanced video patch embeddings. These dynamically generated priors provide precise pixel-level guidance, significantly enhancing the SAM 2's decoding process.

As illustrated in \cref{framework}, we start by establishing inter-frame interaction among the video patch embeddings for each frame. Frame-agnostic video patch embeddings are temporally and spatially unfolded into video embeddings $V^{'}\in \mathbb{R}^{(T\times{N_{v}})\times{D}}$, which are subsequently fed into the multi-head self-attention layer and FFN to model the spatiotemporal context across the $T\times{N_{v}}$ dimension, learning pixel-wise inter-frame global correlations. Afterward, the [CLS] tokens for each frame are similarly flattened across the spatiotemporal dimensions, resulting in video class embeddings $V_{cls}^{'}\in \mathbb{R}^{(T\times{1})\times{D}}$. To generate frame-consistent mask priors while enriching the class embeddings with visual context, we employ a multi-head cross-attention to facilitate spatiotemporal interaction between the video class embedding and the video patch embeddings. Here, the video class embeddings $V_{cls}^{'}$ serve as queries, and the video embeddings $V^{'}$ act as keys and values. The whole process could be formulated as follows:
\begin{equation}
\begin{aligned}
    &V^{'} = \text{R}(V) + \text{MHSA}(\text{R}(V)),\\
    % &V^{'} = R(LN(FFN(LN(V^{'})) + LN(V^{'}))),\\
    &V^{'}_{cls} = \text{R}(V_{cls}) + \text{MHCA}(\text{R}(V_{cls}), V^{'})
\end{aligned}
\end{equation}
where $\text{R}(\cdot)$ denotes the reshape operation, MHSA and MHCA represent multi-head self-attention and cross-attention, respectively.

The preceding steps establish inter-frame coherence in both video patch embeddings $V$ and class embeddings $V_{cls}$. The class embeddings $V_{cls}$, jointly encoded with visual and linguistic modalities, effectively represent the intersection of visual and textual information, specifically the object denoted in the text. Therefore, after the video class embeddings $V_{cls}^{'}$ are broadcast to match the dimensions of the video embeddings $V^{'}$, element-wise multiplication is performed. The resulting embeddings are then processed by a MLP layer to generate the global video feature $V_g$ enriched with object information. Simultaneously, dimensionality reduction is applied to both the video and class embeddings. The video embeddings are reshaped into feature maps with dimensions $\frac{H_m}{p}\times{\frac{W_m}{p}}$. The class embeddings $V_{cls}$, representing foreground information, are multiplied element-wise with these reshaped video embeddings representing global features. This product is subsequently passed through the mask producer, which consists of a MLP layer, to generate the frame-specific mask priors $M_{p}\in{\mathbb{R}^{T\times{\frac{H_m}{p}\times{\frac{W_m}{p}}}}}$. This process could be formulated as follows:
\begin{equation}
V_{g} = \mathrm{MLP}(V^{'}\cdot{V_{cls}^{'}}), M_{p} = \mathrm{MLP}(V_{g}).
\end{equation}

Then, the global video feature $V_{g}$ is input into the hierarchical global-historical aggregator to provide global object context, and the mask priors $M_{p}$ are supplied as dense prompts to the prompt encoder of SAM 2.

\begin{figure}[t]
    \centering
     \includegraphics[width=\linewidth]{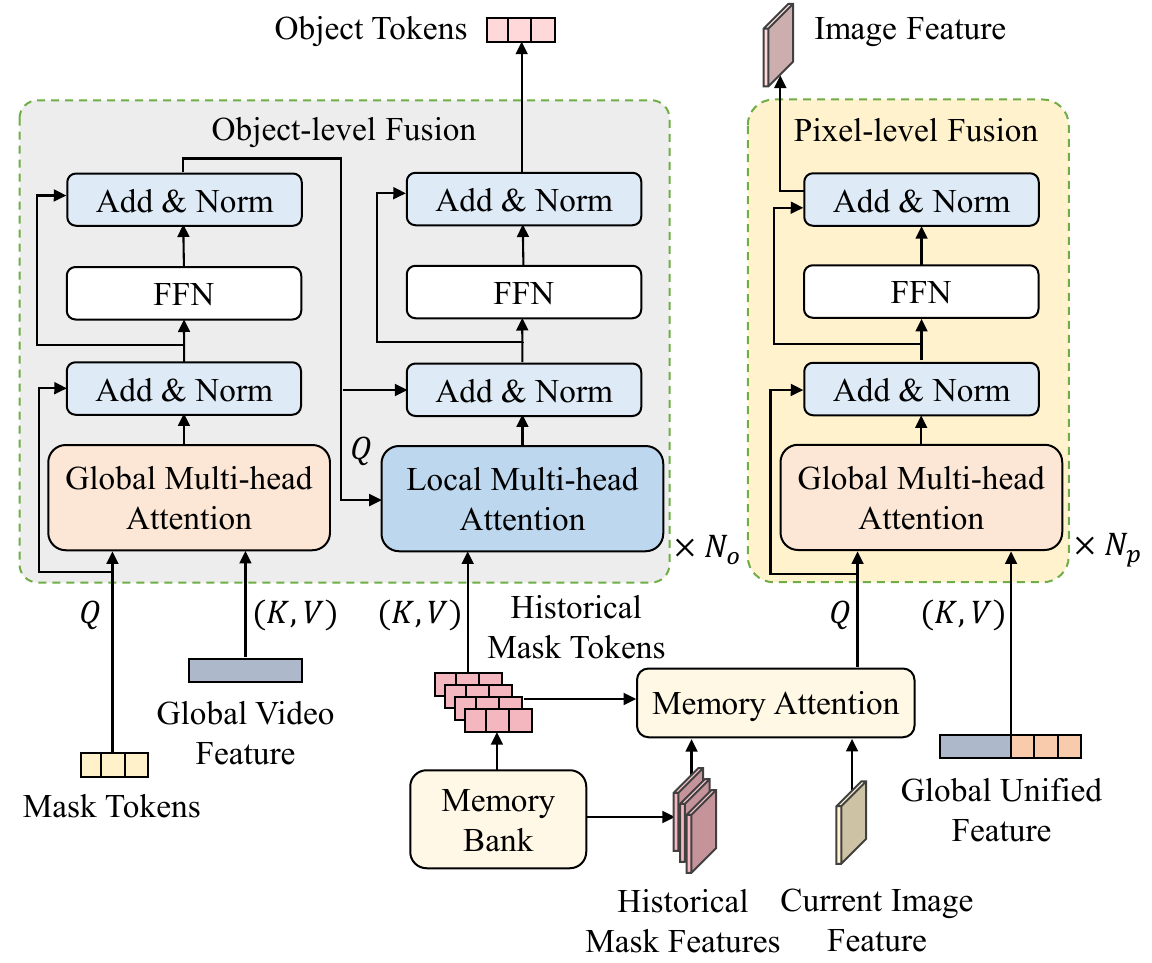}
     \caption{The structure of the hierarchical global-historical aggregator.}
     \label{HGA}
\end{figure} 

\subsection{Hierarchical Global-Historical Aggregator}

Unlike VOS that primarily relies on past information, RVOS emphasizes the effective utilization of both temporal and textual cues. Inspired by \cite{cheng2024putting}, we propose a hierarchical global-historical aggregator that synergistically combines SAM 2's memory mechanism with multi-level temporal modeling, enabling comprehensive integration of global context and historical segmentation results at both pixel-wise and object-centric levels. As shown in \cref{HGA}, the hierarchical global-historical aggregator comprises two components: the pixel-level fusion module and the object-level fusion module.

The pixel-level fusion module processes the current frame's image feature $F_{i}$ through a two-stage attention mechanism. First, memory attention is performed with historical mask features and mask tokens from the memory bank, enhancing the target representation with localized temporal information, resulting in the locally enhanced feature $F_{l}$. Subsequently, $F_{l}$ undergoes global multi-head attention with the global unified features $V_{u}$, which are generated by concatenating compressed global video features $V_{g}$ with corresponding text features. Although text features are encoded independently for each frame, their semantic consistency across the video sequence enables them to serve as global context. The final output of this module, the enhanced frame feature $F_{lg}$, is obtained through the add \& layer norm operation and a FFN layer, effectively combining local and global information.

In the object-level fusion module, the mask tokens of the current frame $T_{m}$ first engage in cross-attention with the global video feature $V_g$ from the mask prior generator. After the add \& layer norm operation and a FFN layer, the globally enhanced mask tokens $T_{mg}$ undergo another cross-attention operation with historical mask tokens from the memory bank. This process yields the final object tokens $T_{mgl}$ that incorporate both global context and historical information through another round of add \& layer norm operation and a FFN layer.

The enhanced object tokens $T_{mgl}$ are then fed into the mask decoder, replacing the original mask tokens, while the enhanced frame features $F_{lg}$ are simultaneously processed. The object queries, composed of sparse prompts and enhanced object tokens, along with the text-enriched frame features $F_{lg}$, collectively enable a more accurate mask decoding process. This dual enhancement strategy ensures that both the object queries and video features are infused with comprehensive temporal and textual information, significantly improving the segmentation accuracy.

\subsection{SAM 2 Prompt Encoder and Mask Decoder}
In MPG-SAM 2, the prompt encoder processes two distinct types of prompts: sparse prompts and dense prompts. For the sparse prompts, following EVF-SAM \cite{zhang2024evf}, we project the [CLS] tokens $V_{cls}$ from the multimodal encoder through a token MLP and then concatenate them with zero-initialized sparse tokens to form the sparse prompts. The dense prompts are generated by upsampling the mask priors $M_{p}$ from the mask prior generator through linear interpolation, ensuring spatial alignment with SAM 2's feature dimensions.

The SAM 2 mask decoder is designed to utilize both prompt types synergistically: (1) the sparse prompts, concatenated with object tokens, form the object-level queries for the decoder, while (2) the dense pixel-level prompts are element-wise added to the video frame features, providing direct guidance from high-resolution feature layers. 

While the original SAM 2 architecture does not inherently support concurrent processing of both prompt types, our modifications enable effective integration of multimodal sparse embeddings and dense pseudo mask priors. This architectural enhancement significantly improves the mask decoder's capability to interpret text-referenced objects, which is crucial for RVOS performance.

\subsection{Training Loss}

MPG-SAM 2 employs an overall loss function similar to that of \cite{Referformer} to constrain the predicted mask as follows:
\begin{equation}
\mathcal{L}=\lambda_{dice}\mathcal{L}_{dice}+\lambda_{focal}\mathcal{L}_{focal}+\lambda_{sim}\mathcal{L}_{sim},
\end{equation}
here, $\mathcal{L}_{dice}$ is the DICE loss \cite{milletari2016v}, $\mathcal{L}_{focal}$ represents the binary mask focal loss and $\mathcal{L}_{sim}$ denotes the mask-text similarity loss, detailed as follows.

\noindent\textbf{Mask-text Similarity Loss.} In the context of the RVOS task, evaluating the similarity between the segmentation mask and the ground truth mask is essential. However, in addition to the traditional mask-based evaluation criteria, a mask-text loss function can also be introduced to assess the segmentation results. The similarity between the text and mask can serve as an additional evaluation metric. Specifically, We use sentence embeddings $T_s$ as the abstract representation of the text embeddings $T$, which is dimensionally compressed to a singular scalar through MLP layers, subsequently expanded to match the dimensions of the mask, and utilized as an output to calculate the pixel-level similarity $S_{tp}$ between the text and the predicted mask $M_{pre}$, as well as the pixel-level similarity $S_{tg}$ between the text and the ground truth mask $M_{gt}$: 
\begin{equation}
\begin{aligned}
    &S_{tp} = f_{cos}(\mathrm{MLP}(T_s), M_{pre}), \\
    &S_{tg} = f_{cos}(\mathrm{MLP}(T_s), M_{gt}), \\
\end{aligned}
\end{equation}
where $f_{cos}$ denotes the cosine similarity function. Subsequently, MSE loss is employed to enforce pixel-level constraints between $S_{tp}$ and $S_{tg}$:
\begin{equation}
    \mathcal{L}_{sim} = \frac{1}{N}\sum_{i=1}^{N}({S_{tp}}_{i}-{S_{tg}}_{i})^2,
\end{equation}
where $N$ represents the number of pixels in the whole video.

\section{Experiments}
\subsection{Datasets and Metrics}
\noindent\textbf{Datasets.} The experiments are performed on several key RVOS datasets: Ref-YouTube-VOS \cite{Urvos}, MeViS \cite{ding2023mevis} and Ref-DAVIS17 \cite{khoreva2018video}. Ref-YouTube-VOS is a widely recognized and large-scale dataset in the field of RVOS, containing 3471 videos with 12913 expressions in the training set and 202 videos with 2096 expressions in the validation set. MeViS, a newly established dataset, focuses on motion analysis, consisting of 2,006 videos with 28,570 annotations. Ref-DAVIS17 builds on the DAVIS17 \cite{pont20172017} dataset with additional linguistic annotations for diverse objects, comprising 90 videos. 

\noindent\textbf{Evaluation Metrics.} We adhere to the standard evaluation framework outlined in \cite{Urvos}, employing metrics such as region similarity $\mathcal{J}$, contour accuracy $\mathcal{F}$, and their combined average $\mathcal{J}\&\mathcal{F}$ to evaluate our model on the validation sets of Ref-Youtube-VOS, MeViS and Ref-DAVIS17. Due to the absence of publicly accessible ground truth annotations for the Ref-Youtube-VOS and MeViS validation set, we utilize the official server to submit our predictions and obtain the evaluation results.

\subsection{Implementation Details}
\noindent\textbf{Model Settings.} We initialize the relevant modules of SAM 2 and the multimodal encoder using the SAM 2-Hiera-Large \cite{ravi2024sam} and BEiT-3-Large \cite{wang2023image} pre-trained weights. For feature parsing, each image is resized to resolutions of $1024\times{1024}$ and $224\times{224}$, serving as the input to SAM 2 image encoder with an output dimension $C$ of 256, and the multimodal encoder with an output dimension $D$ of 1024, respectively. In the hierarchical global-historical aggregator, the patch size $p_{g}$ for the global video feature in pixel-level fusion is set to 2. Both the pixel-level fusion layer number $N_p$  and object-level fusion layer number $N_o$ are set to 1. The memory bank is configured similarly to SAM 2 \cite{ravi2024sam}, with a maximum storage capacity of 7 historical mask features and 16 mask tokens.

\noindent\textbf{Training Details.} Experiments are conducted on 8 NVIDIA A800 GPUs for the MeViS \cite{ding2023mevis} dataset due to its high memory requirements stemming from dataset-specific configuration and on 8 NVIDIA GeForce RTX 4090 GPUs for the remaining datasets. The MeViS dataset experiments follow settings similar to \cite{he2024decoupling}, with 8 frames as input for training, and are trained directly on this dataset without any pre-training on RefCOCO/+/g \cite{mao2016generation, yu2016modeling}. Training is performed for 6 epochs using the AdamW optimizer \cite{loshchilov2018decoupled}, set at a learning rate of 2e-6. For the Ref-YouTube-VOS \cite{Urvos} and Ref-DAVIS17 \cite{khoreva2018video} datasets, we adopt an approach similar to \cite{Referformer, miao2023spectrum}, first pre-training on the RefCOCO/+/g \cite{mao2016generation, yu2016modeling} for 10 epochs, followed by fine-tuning on Ref-YouTube-VOS for 6 epochs. The batch size during pre-training is set to 8, with a learning rate of 1e-5, while fine-tuning use a batch size of 1, a learning rate of 2e-6, and 5 frames as input. To better align with the pre-trained parameters of other modules, we set a learning rate of 5e-5 for both the hierarchical global-historical aggregator and the mask prior generator. The trained model is then validated on the Ref-DAVIS17 dataset without additional training. The loss weights for different losses are set as follows: $\lambda_{focal} = 2$, $\lambda_{dice} = 5$, and $\lambda_{sim} = 2$.

\begin{table}[t]
\setlength{\tabcolsep}{3pt}
\centering
\footnotesize
% \scriptsize
\caption{Comparison with state-of-the-art methods on the Ref-YouTube-VOS and Ref-DAVIS17 datasets. The best results are highlighted in bold, and the second best results are underlined.}
    \begin{center}
        \begin{tabular}{l | c | c c c | c c c}
        \toprule
        \multirow{2}{*}{Method} & \multirow{2}{*}{Reference} & \multicolumn{3}{c|}{Ref-YouTube-VOS} & \multicolumn{3}{c}{Ref-DAVIS17}\\
        & & $\mathcal{J}\&\mathcal{F}$ & $\mathcal{J}$ & $\mathcal{F}$ & $\mathcal{J}\&\mathcal{F}$ & $\mathcal{J}$ & $\mathcal{F}$\\
        \midrule
        ReferFormer \cite{Referformer} & CVPR'22 & 62.9 & 61.3 & 64.6 & 61.1 & 58.1 &64.1 \\
          OnlineRefer \cite{wu2023onlinerefer} & ICCV'23 & 62.9& 61.0& 64.7& 62.4&59.1 &65.6\\
         HTML \cite{han2023html}& ICCV'23 & 63.4 & 61.5 &65.2 &62.1&59.2&65.1\\ 
        SgMg \cite{miao2023spectrum} & ICCV'23 & 65.7 & 63.9 & 67.4 & 63.3&60.6&66.0\\
         TempCD \cite{tang2023temporal} & ICCV'23  & 65.8 & 63.6 & 68.0 & 64.6 & 61.6 & 67.6 \\
        SOC \cite{luo2024soc} & NIPS'23  & 66.0 & 64.1 &67.9 &  64.2&61.0&67.4\\
        LoSh \cite{yuan2024losh} & CVPR'24  & 67.2 & 65.4 & 69.0 & 64.3 & 61.8 & 66.8 \\
        DsHmp \cite{he2024decoupling} & CVPR'24 & 67.1 & 65.0 &69.1 &64.9 & 61.7 &68.1\\
        MUTR \cite{yan2024referred} & AAAI'24 & \underline{68.4} & \underline{66.4}& \underline{70.4} & 68.0 & 64.8 & 71.3 \\
        VD-IT \cite{zhu2024exploring} & ECCV'24 & 66.5 & 64.4 & 68.5 & 69.4 &66.2&72.6 \\
         VISA \cite{yan2024visa} & ECCV'24 &  63.0 &61.4 & 64.7 & \underline{70.4} & \underline{67.0} & \underline{73.8}  \\ 
        \hline
        MPG-SAM 2 & - & \textbf{73.9} & \textbf{71.7} &\textbf{76.1} &\textbf{72.4} &\textbf{68.8} & \textbf{76.0} \\
        \bottomrule
    \end{tabular}
    \end{center}
    \label{Refytvos_RefDavis}
% \vspace{-4mm}
\end{table}

\begin{table}[t]
\setlength{\tabcolsep}{8pt}
\centering
% \small
\footnotesize
\caption{Comparison with state-of-the-art methods on the MeViS dataset. Our model achieves the best performance.}
% \vspace{-4mm}
% \tabcolsep 9pt
    \begin{center}
        \begin{tabular}{l | c | c c c }
        \toprule
        Method & Reference & $\mathcal{J}\&\mathcal{F}$ & $\mathcal{J}$ & $\mathcal{F}$ \\
        \midrule
        URVOS \cite{Urvos} & ECCV'20 & 27.8 & 25.7 & 29.9 \\
        LBDT \cite{ding2022language}& CVPR'22&29.3&27.8&30.8\\
         MTTR \cite{botach2022end}&CVPR'22 &30.0&28.8&31.2\\
         ReferFormer \cite{Referformer} & CVPR'22 &31.0&29.8&32.2\\
         VLT+TC \cite{ding2021vision}& TPAMI'22&35.5&33.6&37.3\\
         LMPM \cite{ding2023mevis} & ICCV'23  &37.2 &34.2& 40.2\\
         VISA \cite{yan2024visa} & ECCV'24 & 44.5&41.8 & 47.1  \\
         DsHmp \cite{he2024decoupling} & CVPR'24 & 46.4& 43.0 &49.8 \\
         \hline
         MPG-SAM 2 & - & \textbf{53.7} & \textbf{50.7} & \textbf{56.7} \\
        \bottomrule
    \end{tabular}
    \end{center}
    \label{Mevis}
\vspace{-2mm}
\end{table}

\subsection{Comparison with State-of-the-Art Methods}

\noindent\textbf{Ref-YouTube-VOS \& Ref-DAVIS17 sets.} We compare our MPG-SAM 2 approach with several state-of-the-art methods, as shown in \cref{Refytvos_RefDavis}. Our approach outperforms all existing methods on both datasets. On the Ref-YouTube-VOS \cite{Urvos} dataset, we achieve 73.9\% $\mathcal{J}\&\mathcal{F}$, surpassing the best methods DsHmp \cite{he2024decoupling} by 6.8\% $\mathcal{J}\&\mathcal{F}$ and LoSh \cite{yuan2024losh} by 6.7\% $\mathcal{J}\&\mathcal{F}$. Even compared to methods using additional training data, our model performs competitively, exceeding MUTR \cite{yan2024referred} by 5.5\% $\mathcal{J}\&\mathcal{F}$. On the Ref-DAVIS \cite{khoreva2018video} dataset, our method achieves 72.4\% $\mathcal{J}\&\mathcal{F}$, outperforming VD-IT \cite{zhu2024exploring} by 3.0\% $\mathcal{J}\&\mathcal{F}$ and surpassing VISA \cite{yan2024visa} by 2.0\% $\mathcal{J}\&\mathcal{F}$, which is trained on additional datasets. All comparison methods use the optimal configuration to highlight model performance.

\noindent\textbf{MeViS set.} We also conduct comparative experiments between our MPG-SAM 2 and existing methods including URVOS \cite{Urvos}, LBDT \cite{ding2022language}, MTTR \cite{botach2022end}, ReferFormer \cite {Referformer}, VLT+TC \cite{ding2021vision}, LMPM \cite{ding2023mevis}, VISA \cite{yan2024visa} and DsHmp \cite{he2024decoupling}, on the MeViS \cite{ding2023mevis} dataset, with results documented in \cref{Mevis}. On this dataset, our method achieves a $\mathcal{J}\&\mathcal{F}$ score of 53.7\%, surpassing the current state-of-the-art method DsHmp by 7.3\% $\mathcal{J}\&\mathcal{F}$, demonstrating the effectiveness of our approach in leveraging temporal information.

\begin{figure*}[t]
\begin{center}
   \includegraphics[width=\linewidth]{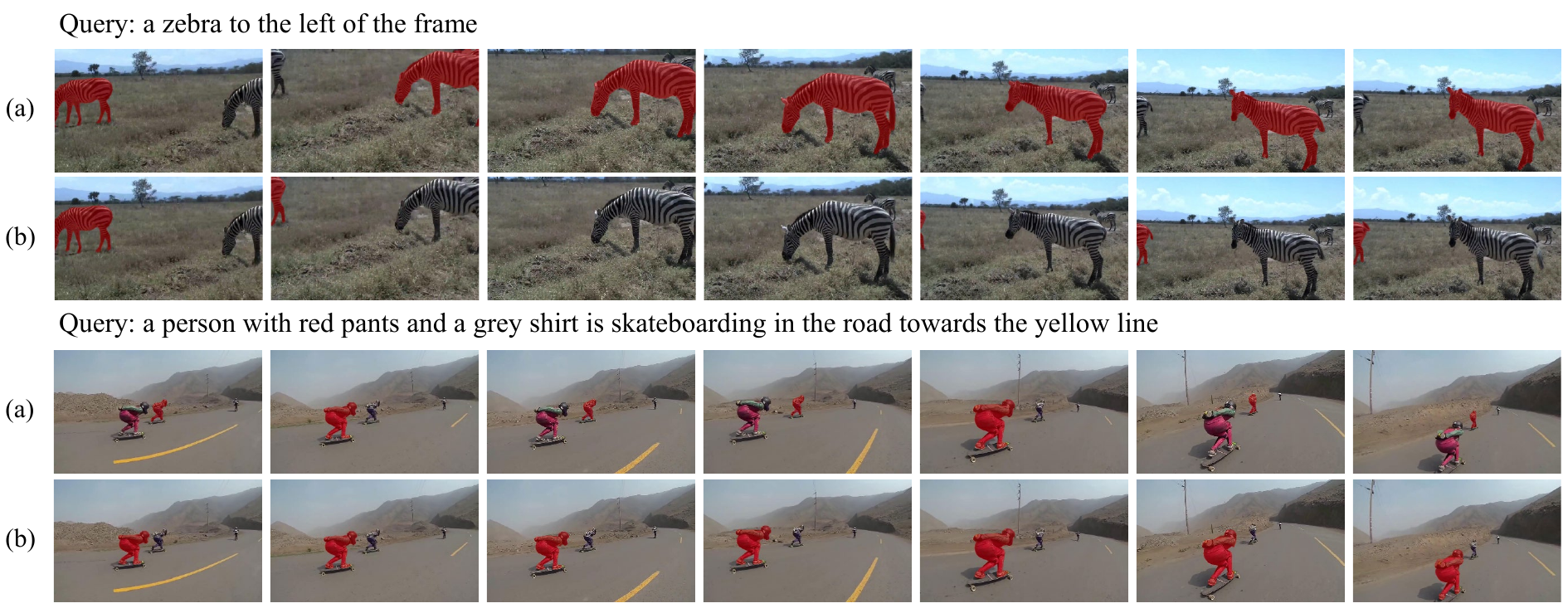}
\end{center}
% \vspace{-4mm}
\caption{Visualization result on Ref-YouTube-VOS. (a) SgMg \cite{miao2023spectrum}, (b) MPG-SAM 2. MPG-SAM 2 effectively ensures the global spatiotemporal consistency of masks, reducing objects misalignment and drift.}
\label{vis_res}
\end{figure*}

\cref{vis_res} presents the visual comparison of our MPG-SAM 2 model with SgMg \cite{miao2023spectrum} on Ref-YouTube-VOS dataset. The results clearly demonstrate that MPG-SAM 2 consistently surpasses SgMg, especially in prediction accuracy and maintaining frame-to-frame consistency. 

Although MPG enables spatiotemporal self-interaction on video embeddings, its computational and memory overhead remains acceptable due to BEiT-3’s small embedding size (14$\times$14). MPG and HGA together add 1.3G to memory and introduce 21M parameters, which is reasonable. For a detailed analysis of the overall model parameters, please refer to the appendix.

\subsection{Model Analysis}
In this section, we conduct comprehensive ablation experiments to examine the effects of the key components of our MPG-SAM 2 and the influence of various model configurations. All experiments are carried out using the Ref-Youtube-VOS dataset.

\begin{table}
\caption{Ablation study of different components of MPG-SAM 2 on Ref-YouTube-VOS dataset.}
\setlength{\tabcolsep}{7pt}
\centering
\footnotesize
% \small
    \begin{center}
        \begin{tabular}{l | c c c| c c c}

\toprule
Method &$\mathcal{L}_{sim}$ &MPG & HGA & $\mathcal{J}\&\mathcal{F}$ & $\mathcal{J}$ & $\mathcal{F}$  \\
\midrule

Baseline & & & & 69.4&67.5&71.3 \\ 
MPG-SAM 2 &\ding{51} & & & 70.3&68.3&72.2 \\ 
MPG-SAM 2 &\ding{51} & \ding{51}& & 71.9&69.8&73.9 \\
MPG-SAM 2 &\ding{51} & &\ding{51}& 72.3&70.1&74.4 \\
MPG-SAM 2 & \ding{51}&\ding{51}&\ding{51}&\textbf{73.9}&\textbf{71.7}&\textbf{76.1} \\

\bottomrule

\end{tabular}
    \end{center}
    \label{components}
\vspace{-2mm}
\end{table}

\noindent\textbf{Components Analysis.} To explore the influence of the key components of our model, we first construct a \textbf{baseline} model consisting solely of SAM 2 and the multimodal encoder. We improve the video baseline over image-based referring segmentation by (1) providing per-frame prompts ($P_{per}$) instead of only the first frame and (2) using the SAM 2 memory mechanism (Mem) instead of frame-independent segmentation. As shown in \cref{ablation}, these enhancements lead to the baseline by 10.4\% $\mathcal{J}\&\mathcal{F}$ and 3.2\% $\mathcal{J}\&\mathcal{F}$, respectively.

After that, as depicted in \cref{components}, we first add the mask-text similarity loss ($\mathcal{L}_{sim}$) to the baseline, resulting in a 0.9\% $\mathcal{J}\&\mathcal{F}$ improvement. Building on this, we introduce the mask prior generator (MPG), and the special MPG-SAM 2 achieves 71.9\% $\mathcal{J}\&\mathcal{F}$, which is 1.6\% higher than the previous model. While only the hierarchical global-historical aggregator (HGA) is imposed on the baseline with $\mathcal{L}_{sim}$, the $\mathcal{J}\&\mathcal{F}$ score of the special MPG-SAM 2 reaches 72.3\%, indicating a 2.0\% improvement and further validating the module's superiority. With all components integrated, our MPG-SAM 2 delivers the best performance of 73.9\% $\mathcal{J}\&\mathcal{F}$. Note that the memory mechanism is excluded from HGA during performance calculations; when the MPG module is omitted, only the mask prior is not generated, while the global video feature production remains.

\begin{table}
\caption{Model analysis of different settings in MPG-SAM 2.}
% \vspace{-4mm}
\setlength{\tabcolsep}{9pt}
\centering
% \small
\footnotesize
    \begin{center}
        \begin{tabular}{l | c | c c c}

\toprule

{Method} & {Settings} &$\mathcal{J}\&\mathcal{F}$ & $\mathcal{J}$ & $\mathcal{F}$ \\

\hline
\multicolumn{5}{l}{\textbf{Settings of Baseline}} \\
\hline
Baseline& w/o $P_{per}$ & 59.0 & 57.5 & 60.6 \\
Baseline & w/o Mem & 66.2 & 64.5 & 67.9  \\
Baseline & w $P_{per}$ \& Mem & \textbf{69.4} & \textbf{67.5} & \textbf{71.3} \\ 
\hline
\multicolumn{5}{l}{\textbf{Interaction Mode of MPG}} \\
\hline
MPG-SAM 2& w/o $S_{si}$ & 73.3 & 71.2 & 75.4 \\
MPG-SAM 2 & w/o $S_{ci}$ & 73.5 & 71.4 & 75.6  \\
MPG-SAM 2 & w $S_{si}$ \& $S_{ci}$ & \textbf{73.9} & \textbf{71.7} & \textbf{76.1} \\ 
\hline
\multicolumn{5}{l}{\textbf{Patch Size of HGA}} \\
\hline
MPG-SAM 2 & 1 &73.2&71.0&75.4\\
MPG-SAM 2 & 2 & \textbf{73.9} & \textbf{71.7} & \textbf{76.1}   \\
MPG-SAM 2 & 4 & 73.6& 71.4&75.7\\ 
\hline
\multicolumn{5}{l}{\textbf{Global Video Feature of HGA}} \\
\hline
MPG-SAM 2 & Vanilla & 73.1 & 71.1& 75.1 \\
MPG-SAM 2 & Masked & \textbf{73.9} & \textbf{71.7} & \textbf{76.1}  \\
\bottomrule

\end{tabular}
    \end{center}
    \label{ablation}
\vspace{-4mm}
\end{table}

\noindent\textbf{Mask Prior Generator.} In this section, we investigate different spatiotemporal interaction forms within the mask prior generator (MPG), with results presented in \cref{ablation}. When the overall spatiotemporal self-interaction ($S_{si}$) of video embeddings is omitted, MPG-SAM 2 experiences a performance drop of 0.6\% $\mathcal{J}\&\mathcal{F}$. Similarly, when the spatiotemporal cross-modal interaction ($S_{ci}$) between the [CLS] tokens and video embeddings is excluded, the model performance decreases by 0.4\% $\mathcal{J}\&\mathcal{F}$. These findings highlight the importance of understanding video information from a global spatiotemporal perspective during the mask prompt generation process.

\noindent\textbf{Hierarchical Global-Historical Aggregator.} The impact of different settings of hierarchical global-historical aggregator (HGA) is also worth noting. First, we examine the effect of patch size $p_{g}$ in the pixel-level fusion process for the global video feature, experimenting with patch sizes 1, 2, and 4. As shown in \cref{ablation}, the patch size of 2 yields the best model performance. This configuration achieves an optimal balance between critical and redundant information during pixel-level fusion, allowing for more effective integration of global context.

Additionally, we explore different configurations for the global video feature input to the hierarchical global-historical aggregator. One straightforward approach named ``Vanilla" employs the video embeddings directly output by the multimodal encoder as the global feature. An alternative approach dubbed ``Masked" leverages a video feature enriched with mask prior information. Results in \cref{ablation} indicate that video features containing mask information are more beneficial for global context integration, as they better emphasize global frame mask information to guide the segmentation of the current frame.

\section{Conclusion}
In this paper, we present MPG-SAM 2, an innovative end-to-end framework for RVOS, to address the challenges of adapting SAM 2 to the RVOS task. Our approach utilizes a unified multimodal encoder to jointly encode video and textual features, generating semantically aligned video and text embeddings, along with multimodal class tokens. The video embeddings and class tokens are employed by a mask prior generator to create pseudo masks of target objects, offering strong positional cues as dense prompts for SAM 2's mask decoder. To address SAM 2’s lack of global context awareness in offline RVOS, we introduce a hierarchical global-historical aggregator. This enables SAM 2 to integrate global context and historical information of target objects at both pixel and object levels, enhancing the target representation and temporal consistency. Extensive experiments on several RVOS benchmarks demonstrate the superiority of our MPG-SAM 2 over state-of-the-art methods and validate the effectiveness of our proposed modules.

\appendix
\clearpage
\twocolumn[
\begin{center}
\Large\bfseries Supplementary Material\\
\end{center}
]
\section{Additional Experimental Studies}
\noindent\textbf{Mask-text Similarity Loss.} In this part, We validate the generalizability of the mask-text similarity loss function by conducting enhancement experiments with this function on several previous RVOS methods, including ReferFormer \cite{Referformer} and SgMg \cite{miao2023spectrum}. Meanwhile, we also validate the impact of this function on the performance of MPG-SAM 2. The experimental results, presented in \cref{loss}, indicate performance improvements across all methods, confirming the effectiveness of the proposed similarity function in RVOS tasks.

\noindent\textbf{Model Parameters.} In this section, we analyze the parameter count of our model. We supplement our study with a set of low-configuration experiments on the Ref-YouTube-VOS \cite{Urvos} dataset, using SAM 2-Hiera-Large \cite{ravi2024sam} and BEiT-3-Base \cite{wang2023image} as initialization parameters, called MPG-SAM 2-Tiny. The experimental results and parameter counts are presented in \cref{Param}. Compared to previous methods with relatively small parameter sizes, such as ReferFormer \cite{Referformer} and SgMg \cite{miao2023spectrum}, our low-configuration model MPG-SAM 2-Tiny exhibits a slightly larger parameter count but achieves a substantial performance gain. Furthermore, although methods like VISA \cite{yan2024visa} and HyperSeg \cite{wei2024hyperseg} also employ vision-language models and are trained on additional datasets, in contrast, our full-configuration model MPG-SAM 2, which is not trained on any additional datasets, demonstrates superior performance with a smaller model size. This highlights the effectiveness and efficiency of our approach.

\noindent\textbf{Hierarchical Global-Historical Aggregator.} In this division, we perform a more detailed component analysis to evaluate the effectiveness of each part within the hierarchical global-historical aggregator (HGA) on the Ref-YouTube-VOS \cite{Urvos} dataset. The experimental results are presented in \cref{HGA}. The initial setup involves the MPG-SAM 2 only using the mask prior generator, which achieves 71.9\% $\mathcal{J}\&\mathcal{F}$. When incorporating the global enhancement and local enhancement of HGA's object-level fusion part separately, the model reaches $\mathcal{J}\&\mathcal{F}$ scores of 72.2\% and 72.4\%, respectively, resulting in gains of 0.3\% and 0.5\% $\mathcal{J}\&\mathcal{F}$. When applying both components of the object-level fusion part to the initial model simultaneously, the model attains 72.8\% $\mathcal{J}\&\mathcal{F}$, indicating a 0.9\% $\mathcal{J}\&\mathcal{F}$ score improvement compared to the initial setup. Moreover, the inclusion of HGA's pixel-level global fusion part on the initial setup obtains a $\mathcal{J}\&\mathcal{F}$ score of 73.1\%. Finally, when all parts of HGA are employed, the full model gains the highest performance of 73.9\% $\mathcal{J}\&\mathcal{F}$.

\begin{table}
\caption{Generalizability of the mask-text similarity loss.}

\setlength{\tabcolsep}{4pt}
\centering
\footnotesize
    \begin{center}
        \begin{tabular}{l | c | c c c}

\toprule
Method & Backbone & $\mathcal{J}\&\mathcal{F}$ & $\mathcal{J}$ & $\mathcal{F}$  \\
\midrule
ReferFormer \cite{Referformer} & ResNet-50 & 55.6 & 54.8 &56.5 \\
ReferFormer \cite{Referformer} + $\mathcal{L}_{sim}$ & ResNet-50 & \textbf{56.4} & \textbf{55.4} &\textbf{57.5} \\
\hline
SgMg \cite{miao2023spectrum} & Video-Swin-T & 62.0 & 60.4 & 63.5 \\
SgMg \cite{miao2023spectrum} + $\mathcal{L}_{sim}$ & Video-Swin-T & \textbf{62.8} & \textbf{61.3} & \textbf{64.3} \\
\hline
MPG-SAM 2 - $\mathcal{L}_{sim}$ & Hiera-L & 73.2 & 71.2 & 75.2\\
MPG-SAM 2 & Hiera-L & \textbf{73.9} & \textbf{71.7} & \textbf{76.1} \\
\bottomrule

\end{tabular}
    \end{center}
    \label{loss}

\end{table}

\begin{table}
\caption{Model parameter analysis on Ref-YouTube-VOS dataset. Our model strikes a balance between the number of parameters and performance, demonstrating clear advantages. The best results are high-lighted in bold, and the second best results are underlined.}

\setlength{\tabcolsep}{3pt}
\centering
\footnotesize
    \begin{center}
        \begin{tabular}{l | c | c | c c c }
        \toprule
        Method & Reference & All Params & $\mathcal{J}\&\mathcal{F}$ & $\mathcal{J}$ & $\mathcal{F}$ \\
        \midrule
         ReferFormer \cite{Referformer} & CVPR'22 & 0.24B &62.9 & 61.3 & 64.6\\
         SgMg \cite{miao2023spectrum} & ICCV'23 & 0.24B & 65.7 & 63.9 & 67.4 \\
         VISA \cite{yan2024visa} & ECCV'24 & 13B & 63.0  &61.4 & 64.7 \\
        HyperSeg \cite{wei2024hyperseg} & Arxiv'24 & 3B  & 68.5 & - & - \\ 
         \hline
         MPG-SAM 2-Tiny & - & 0.46B &\underline{69.9} & \underline{68.0} & \underline{71.8} \\
           MPG-SAM 2 & - & 0.92B& \textbf{73.9} & \textbf{71.7} &\textbf{76.1}\\
        \bottomrule
    \end{tabular}
    \end{center}
    \label{Param}

\end{table}

We also analyze the effect of the number of the pixel-level fusion layer $N_p$ and object-level fusion layer $N_o$ on the model's performance. For $N_p$ and $N_o$, we design experiments with 1, 2, and 3 layers, and the results are shown in \cref{layer}. The results show that single-layer fusion modules are sufficient to effectively enhance global and historical information at both the pixel level and object level. However, increasing the number of layers introduces redundant information, which may impair the segmentation process for the current frame. As a result, we set both the $N_p$ and $N_o$ to 1 to ensure optimal model performance.

\begin{table}[t]
\caption{The ablation experiments of HGA components, where OGF represents the object-level global fusion part, OLF denotes the object-level local fusion part and PGF refers to the pixel-level global fusion part of HGA.}
\footnotesize
\tabcolsep 4pt
    \begin{center}
        \begin{tabular}{l|c c c | c c c}

\toprule
Method & OGF & OLF & PGF & $\mathcal{J}\&\mathcal{F}$ & $\mathcal{J}$ & $\mathcal{F}$   \\
\midrule

MPG-SAM 2 & & & & 71.9 & 69.8 & 73.9  \\
MPG-SAM 2 & \checkmark & & & 72.2& 70.1 & 74.3\\
MPG-SAM 2 & & \checkmark & & 72.4& 70.1 & 74.6 \\
MPG-SAM 2 & \checkmark & \checkmark & & 72.8 & 70.7 & 75.0 \\
MPG-SAM 2 & & & \checkmark & 73.1 & 71.0 & 75.3 \\
MPG-SAM 2 & \checkmark & \checkmark & \checkmark & \textbf{73.9} & \textbf{71.7} & \textbf{76.1} \\

\bottomrule

\end{tabular}
    \end{center}
    \label{HGA}
\end{table}

\begin{table}[t]
\caption{Performance analysis of the hierarchical global-historical aggregator with varying numbers of layer.}
\small
\tabcolsep 5pt
    \begin{center}
        \begin{tabular}{l | c | c c c}

\toprule

{Method} & {Settings} & $\mathcal{J}\&\mathcal{F}$ & $\mathcal{J}$ & $\mathcal{F}$  \\
\midrule
MPG-SAM 2 & $N_{p}=1$ & \textbf{73.9} & \textbf{71.7} &\textbf{76.1} \\
MPG-SAM 2 & $N_{p}=2$ & 73.6 & 71.6 & 75.7\\
MPG-SAM 2 & $N_{p}=3$ & 73.0 & 70.9 & 75.2 \\ 
\hline

MPG-SAM 2 & $N_{o}=1$ & \textbf{73.9} & \textbf{71.7} &\textbf{76.1} \\
MPG-SAM 2 & $N_{o}=2$ & 73.4 & 71.2 & 75.6\\
MPG-SAM 2 & $N_{o}=3$ & 72.9 & 70.7 & 75.1 \\ 
\bottomrule

\end{tabular}
    \end{center}
    \label{layer}
\end{table}

\begin{figure*}[t]
\begin{center}
   \includegraphics[width=\linewidth]{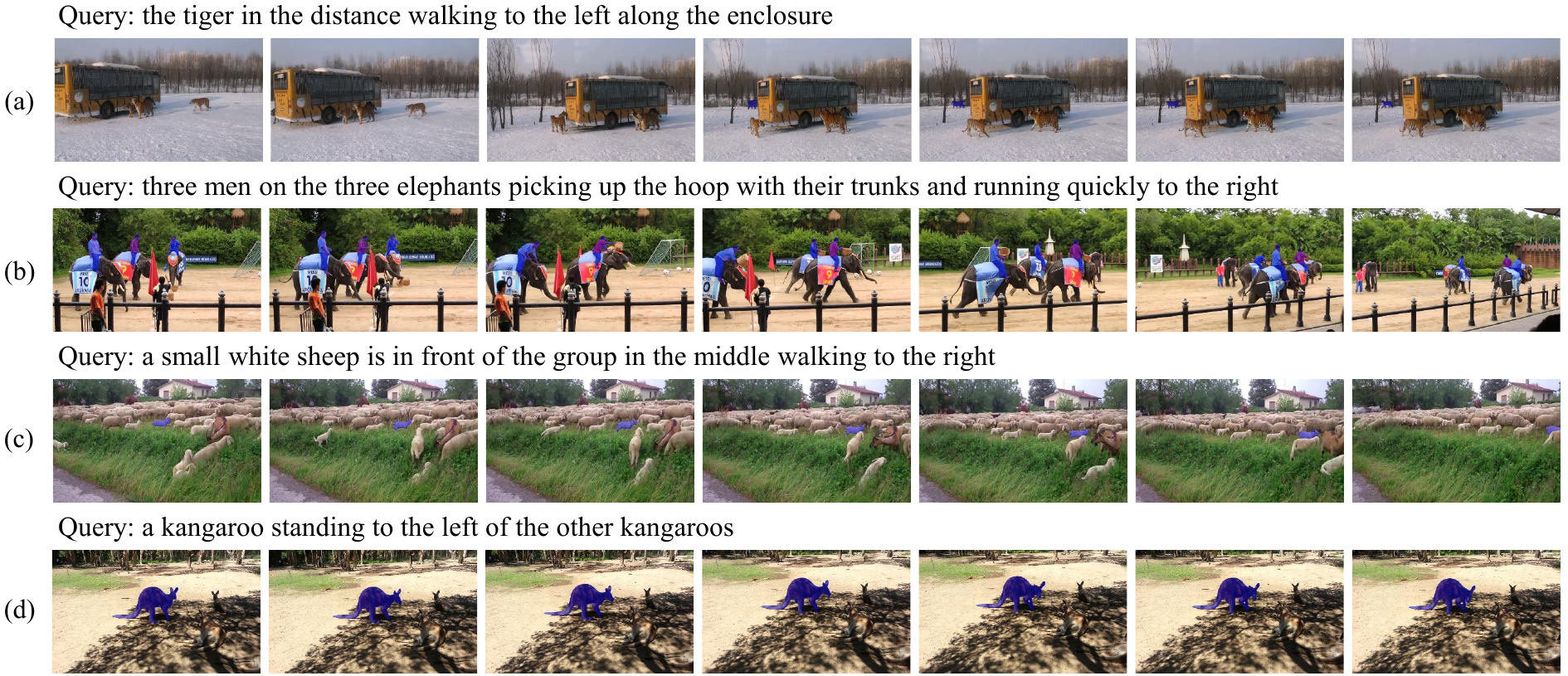}
\end{center}
\caption{Additional visualization results on several datasets. (a), (b) MeViS, (c), (d) Ref-YouTube-VOS.}
\label{vis_sup}
\end{figure*}

\section{Additional Visualization Results}
In this section, we present additional visualization results of our MPG-SAM 2 on Ref-YouTube-VOS \cite{Urvos} dataset and MeViS \cite{ding2023mevis} dataset. The visualizations, shown in \cref{vis_sup}, highlight target objects covered by blue masks.

For the MeViS \cite{ding2023mevis} dataset, we evaluate the model's segmentation performance in challenging scenarios involving object occlusion and multiple referential targets. As shown in \cref{vis_sup} (a), the target object, a tiger, is initially occluded by an enclosure in the first three frames and becomes visible in the subsequent four frames. The model effectively detects the absence of the target in the occluded frames and accurately segments it once it appears. Additionally, \cref{vis_sup} (b) presents a scene with three distinct referential targets, all of which are precisely segmented by the model without any omissions across frames.

For the Ref-YouTube-VOS \cite{Urvos} dataset, we select several complex scenarios featuring multiple similar objects to evaluate the model's ability to distinguish between such targets. As shown in \cref{vis_sup} (c), the model accurately segments the specific sheep matching the language description from a group of visually similar sheep. In \cref{vis_sup} (d), the model successfully disregards the interference caused by shadows and accurately segments the kangaroo positioned on the left. These findings highlight the robustness and effectiveness of our model in tackling diverse and challenging scenarios.

\section*{Acknowledgments}
This work was supported in part by the National Natural Science Foundation of China under Grant 62431020, in part by the Foundation for Innovative Research Groups of Hubei Province under Grant 2024AFA017, in part by the Fundamental Research Funds for the Central Universities under Grant 2042025kf0030.

\clearpage
{
    \small
    \bibliographystyle{ieeenat_fullname}
    \bibliography{main}
}

\end{document}